\def\year{2020}\relax
\documentclass[letterpaper]{article} 
\usepackage{aaai20}  
\usepackage{times}  
\usepackage{helvet} 
\usepackage{courier}  
\usepackage[hyphens]{url}  
\usepackage{graphicx} 
\usepackage{graphicx}  
\usepackage{amsmath}
\usepackage{amsthm}
\usepackage{amssymb,amsfonts}
\usepackage{makecell}
\usepackage{enumerate}
\usepackage{algorithm}
\usepackage{algorithmic}
\usepackage{array}
\usepackage{booktabs}
\usepackage{multirow}

\newcommand{\argmin}{\operatornamewithlimits{arg\,min}}

\newtheorem{lemma}{Lemma}
\newtheorem{theorem}{Theorem}

\newtheorem{definition}{Definition}
\newtheorem{remark}{Remark}
\newtheorem{proposition}{Proposition}
\def\WW{\boldsymbol W}
\def\OO{\boldsymbol \Omega}

\def\xx{\boldsymbol x}
\def\yy{\boldsymbol y}

\def\oo{\boldsymbol \omega}

\title{Automated Spectral Kernel Learning}

\author{
Jian Li,\textsuperscript{\rm 1,2}
Yong Liu,\thanks{Corresponding author}\textsuperscript{\rm 1,2}
Weiping Wang\textsuperscript{\rm 1}\\
\textsuperscript{\rm 1}Institute of Information Engineering, Chinese Academy of Sciences\\
\textsuperscript{\rm 2}School of Cyber Security, University of Chinese Academy of Sciences\\
\{lijian9026, liuyong, wangweiping\}@iie.ac.cn
}

\begin{document}

\maketitle

\begin{abstract}
    The generalization performance of kernel methods is largely determined by the kernel, but spectral representations of stationary kernels are both input-independent and output-independent, which limits their applications on complicated tasks. In this paper, we propose an efficient learning framework that incorporates the process of finding suitable kernels and model training. Using non-stationary spectral kernels and backpropagation w.r.t. the objective, we obtain favorable spectral representations that depends on both inputs and outputs. Further, based on Rademacher complexity, we derive data-dependent generalization error bounds, where we investigate the effect of those factors and introduce regularization terms to improve the performance. Extensive experimental results validate the effectiveness of the proposed algorithm and coincide with our theoretical findings.
\end{abstract}

\section{Introduction}
Kernel methods have achieved great success in many conventional domains over past decades, while they show relatively inferior performance on complicated tasks nowadays.
The fundamental limitation of common kernels has been revealed that they are both \textit{stationary} and \textit{monotony} \cite{bengio2006curse}.
The \textit{stationary} property shows that stationary kernels only depend on the distance $\|\xx - \xx'\|$ while are free from the input $\xx$ itself.
The \textit{monotony} property indicates that values of stationary kernels decrease over the distance, ignoring the long-range interdependence.

Spectral approaches were developed to fully characterize stationary kernels with concise representation forms, such as sparse spectrum kernels \cite{quia2010sparse}, sparse mixture kernels \cite{wilson2013gaussian} and random Fourier features methods to handle with large scale settings \cite{rahimi2007random,le2013fastfood,li2019distributed}.
With sound theoretical guarantees, namely Bochner's theorem \cite{rudin1962fourier,stein2012interpolation}, spectral kernels are constructed from the inverse Fourier transform in the frequency domain.
Although approximate spectral representation provides an efficient approach for stationary spectral kernels, the performance of random features is limited because stationary kernels are \textit{input-independent} and \textit{output-independent}.
Yaglom's theorem, rather than Bochner's theorem, provides more general forms which encompass both stationary and non-stationary kernels via inverse Fourier transform \cite{yaglom1987correlation,samo2015generalized}.
Recently, due to its general and concise spectral statements, non-stationary kernels have been applied to Gaussian process regression \cite{remes2017non,sun2019functional}.

Hyper parameters for kernels determine the performance of kernel methods.
Meanwhile, kernel selection approaches have been well-studied \cite{liu2014kernel,li2017efficient,ding2018randomized,liu2018fast,liu2019fast}.
However, due to the separation of kernel selection and model training, those techniques are inefficient and lead the undesirable performance.

\begin{figure}[t]
    \begin{center}
        \includegraphics[width=.79\columnwidth]{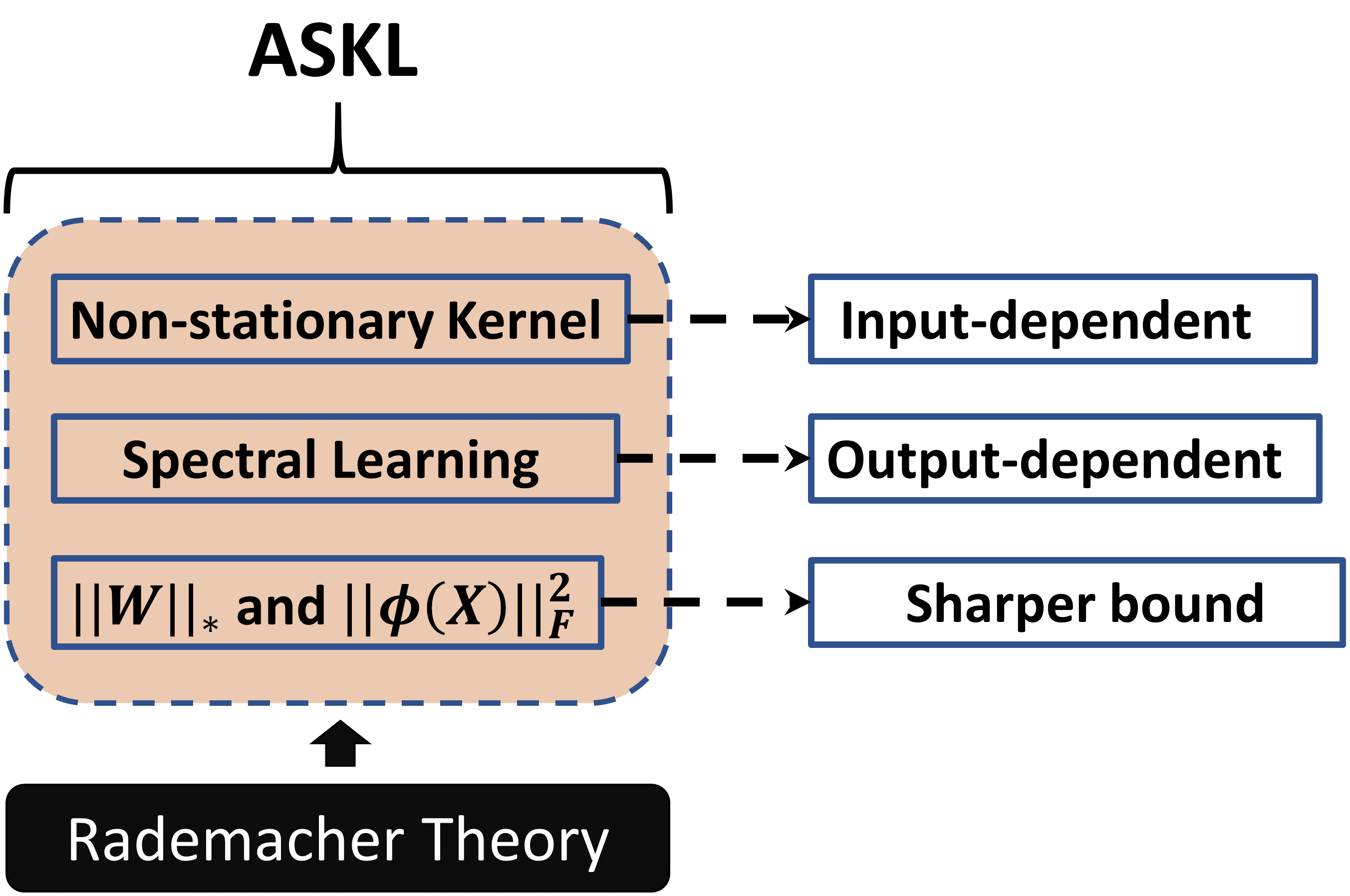}
    \end{center}
    \caption{The overview of \texttt{ASKL}.}
    \label{fig.core_idea}
\end{figure}

In this paper, to achieve better performance ability for kernel methods, we propose an efficient algorithm, namely Automated Spectral Kernel Learning (\texttt{ASKL}), learning suitable kernels and model weights together.
A brief overview is illustrated in Figure \ref{fig.core_idea}.
On the algorithmic front, \texttt{ASKL} consists of:
(1) non-stationary kernels to obtain input-dependent features,
(2) backpropagation w.r.t the objective to make features output-dependent,
(3) regularization terms to achieve sharper generalization error bounds.
On the theoretical front, the theoretical underpinning is Rademacher complexity theory, which indicates how the feature mappings affect the performance and suggests ways to refine the algorithm.

\section{Background}

In ordinary supervised learning settings, training samples $\{(\xx_i, \yy_i)_{i=1}^n\}$ are drawn i.i.d. from a fixed but unknown distribution $\rho$ over $\mathcal{X} \times \mathcal{Y}$, where $\mathcal{X} = \mathbb{R}^d$ is the input space and $\mathcal{Y} \subseteq \mathbb{R}^K$ is the output space in single-valued ($K = 1$) or vector-valued ($K > 1$) forms.
The goal is to learn an estimator $f: \mathcal{X} \to \mathcal{Y}$, which outputs $K$ predictive labels.
We define a standard hypothesis space for kernel methods
\begin{align*}
    \mathcal{H} = \left\{\xx \to f(\xx)=\WW^T\phi(\xx)\right\},
\end{align*}
where $\WW \in \mathbb{R}^{D \times K}$ is the model weight, $\phi(\xx): \mathbb{R}^d \to \mathbb{R}^D$ is a nonlinear feature mapping.
For kernel methods, $\phi(\xx)$ is an implicit feature mapping associated with a Mercer kernel $k(\xx, \xx') = \langle \phi(\xx), \phi(\xx')\rangle$.
To improve the computational efficiency but also retain favorable statistical properties, random Fourier features were proposed to approximate kernel with explicit feature mappings $\phi(\xx)$ via $k(\xx, \xx') \approx \langle \phi(\xx), \phi(\xx')\rangle$ \cite{rahimi2007random}.

In statistical learning theory, the supervised learning problem is to minimize the expected loss on $\mathcal{X} \times \mathcal{Y}$
\begin{align}
    \label{equation.expected-loss}
    \inf_{f \in \mathcal{H}} \mathcal{E}(f), \quad \mathcal{E}(f) = \int_{\mathcal{X} \times \mathcal{Y}} \ell(f(\xx), \yy) ~ d \rho(\xx,\yy),
\end{align}
where $\ell$ is a loss function related to specific tasks.

\subsection{Stationary Kernels}
The connection between the stationary kernel $k(\tau)$ and its spectral density $s(\oo)$ is revealed in Bochner's theorem via inverse Fourier transform.
\begin{theorem}[Bochner's theorem]
    A stationary continuous kernel $k(\xx, \xx') = k(\xx - \xx')$ on $\mathbb{R}^d$ is positive
    definite if and only if it can be represented as
    \begin{align}
        \label{equation.stationary}
        k(\xx, \xx') & = \int_{\mathbb{R}^d} e^{i \oo^T(\xx - \xx')} s(\oo) d \oo,
    \end{align}
    where $s(\oo)$ is a non-negative measure.
\end{theorem}
The spectral density $s(\oo)$ is the probability density of the corresponding kernel.
From the inverse Fourier transform, we find that spectral kernels are highly relevant to the probability measure $s(\oo)$.
E.g. $s(\oo)$ for Gaussian kernels with the width $\sigma$ correspond to Gaussian distribution $\mathcal{N}(0, 1/\sigma)$.

\subsection{Non-Stationary Kernels}
While the \textit{stationary} kernels ignore input-dependent information and long-range correlations, non-stationary kernels alleviate these restrictions because they depend on the inputs themselves \cite{samo2015generalized}.
\begin{theorem}[Yaglom's theorem]
    A general kernel $k(\xx, \xx')$ is positive definite on $\mathbb{R}^d$ is positive define if and only if it admits the form
    \begin{align}
        \label{equation.yaglom}
        k(\xx, \xx') & = \int_{\mathbb{R}^d \times \mathbb{R}^d} e^{i (\oo^T\xx - \oo'^T\xx')} \mu (d \oo, d \oo'),
    \end{align}
    where $\mu (d \oo, d \oo')$ is the Lebesgue-Stieltjes measure associated to some positive semi-definite (PSD) spectral density function $s(\oo, \oo')$ with bounded variations.
\end{theorem}
When $\mu$ is concentrated on the diagonal $\oo = \oo'$, the spectral characterization of stationary kernels in the Bochner's theorem is recovered.
$s(\oo, \oo')$ is a joint probability density.


\section{Automated Spectral Kernel Learning}
In this section, we devise a learning framework for arbitrary kernel-based supervised applications:
\begin{itemize}
    \item We present the learning framework with the minimization objective, integrating empirical risk minimization (ERM) with regularizers on feature mappings and model weights.
    \item Spectral representation for non-stationary spectral kernels based on Yaglom's theorem is conducted.
    \item We apply first-order gradient approaches to solve the minimization objective.
          We update frequency matrices $\OO, \OO'$ together with model weights $\WW$ via backpropagation.
\end{itemize}

\subsection{Learning Framework}
\begin{figure}[t]
    \centering
    \includegraphics[width=.91\columnwidth]{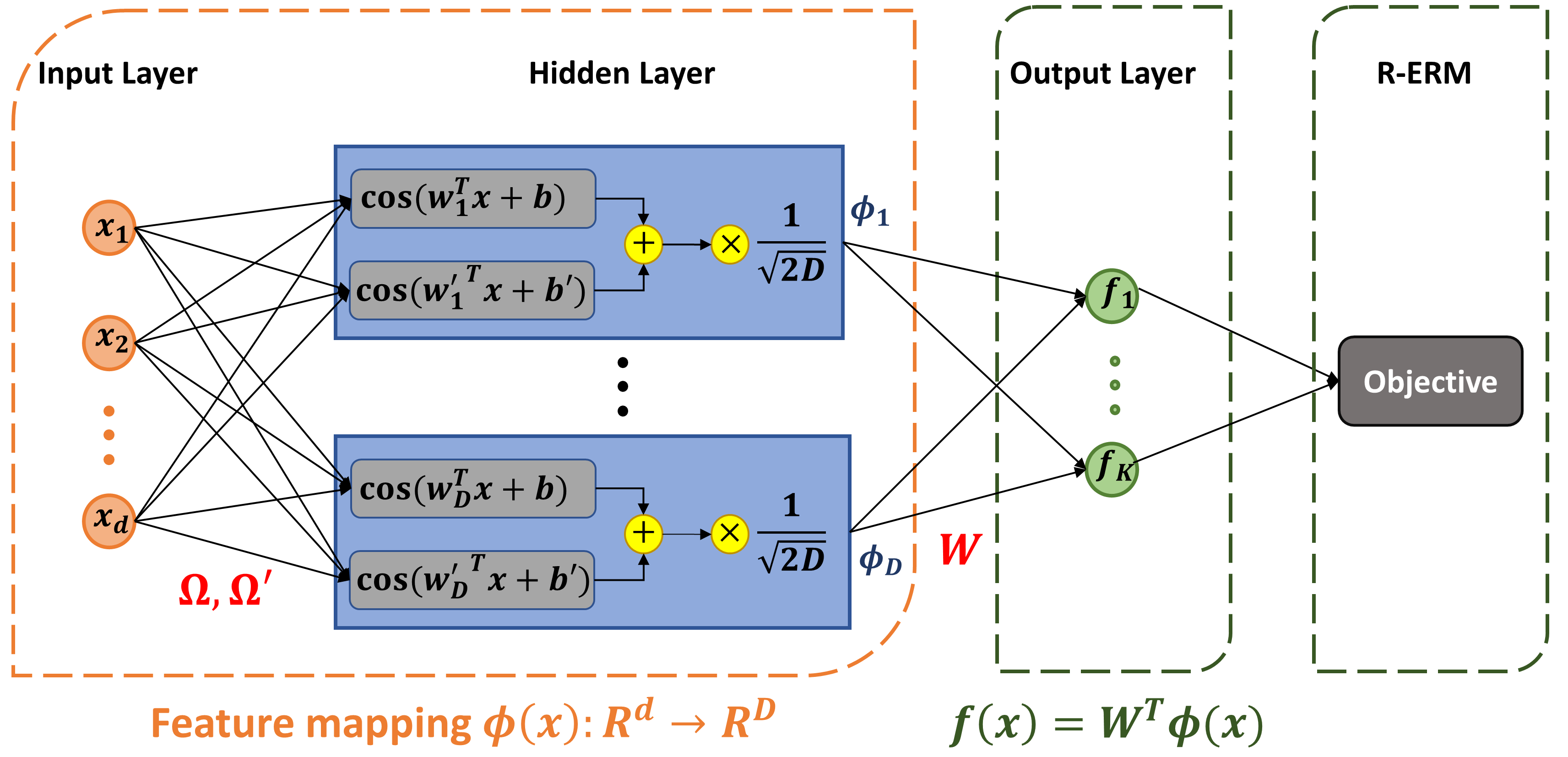}
    \caption{The architecture of learning framework}
    \label{fig.architecture}
\end{figure}
The minimization of the expected loss (\ref{equation.expected-loss}) is hard to estimate in practical problems.
In this paper, we put two additional regularization terms into the ERM, thus the empirical objective $\widehat{\mathcal{E}}(\WW, \OO, \OO')$ becomes
\begin{equation}
    \label{equation.primal-objective}
    \resizebox{0.89\linewidth}{!}{$
    \displaystyle
    \begin{split}
        \argmin_{\WW, \OO, \OO'} \underbrace{\frac{1}{n} \sum_{i=1}^n \ell(f(\xx_i), \yy_i)}_{g(\WW)} + \lambda_1 \|\WW\|_* + \lambda_2 \|\phi(\boldsymbol{X})\|_F^2,
    \end{split}
    $}
\end{equation}
where both feature mappings $\phi(\boldsymbol{X}) \in \mathbb{R}^{D \times n}$ on all data and $f(\xx_i)=\WW^T\phi(\xx_i) \in \mathbb{R}^D$ use the non-stationary spectral representation $\phi(\cdot)$, which is presented as
\begin{align*}
    \phi(\xx) = \frac{1}{\sqrt{2D}}
    \begin{bmatrix}
        \cos(\OO^T \xx + {\boldsymbol b}) + \cos(\OO'^T \xx + {\boldsymbol b}')
    \end{bmatrix}.
\end{align*}
The trace norm $\|\WW\|_*$ and the squared Frobenius norm $\|\phi(\boldsymbol{X})\|_F^2$ exerts constraints on updating model weights $\WW$ and frequency matrices $\OO, \OO'$, respectively.
Those two regularization terms, especially the norm on feature mappings, are rarely used in the R-ERM, leading to the better generalization performance as proven in next section.

In the objective (\ref{equation.primal-objective}), we update model weights $\WW$ and frequency pairs $\OO, \OO'$ to learn feature mappings both input-dependent (non-stationary kernels) and output-dependent (backpropagation towards the objective).
As shown in Figure \ref{fig.architecture}, the architecture can be regarded as a single hidden layer neural network with cosine as activation. 
The spectral density surface $s(\oo, \oo')$ (joint probability density), determining the performance of spectral kernels, is optimized by updating frequency matrices $\OO, \OO'$ via backpropagation \cite{rumelhart1988learning}.
In this structure, only $\WW$ and $\OO, \OO'$ are trainable and optimized towards the objective.

\subsection{Non-Stationary Spectral Kernels Representation}
According to Yaglom's theorem, to produce a positive semi-definite (PSD) kernel, the spectral density needs to be a PSD function.
In order to construct a PSD spectral density $s(\oo, \oo')$, we include symmetries $s(\oo, \oo') = s(\oo', \oo)$ and sufficient diagonal components $s(\oo, \oo)$ and $s(\oo', \oo')$.
We integrate exponential components and the corresponding integrated spectral density $s(\oo, \oo')$ and get
\begin{align*}
    \mathcal{E}_{\oo, \oo'}(\xx, \xx')= \frac{1}{4} \big[
        e^{i (\oo^T\xx - \oo'^T\xx')} + e^{i (\oo'^T\xx - \oo^T\xx')} \\
        + e^{i (\oo^T\xx - \oo^T\xx')} + e^{i (\oo'^T\xx - \oo'^T\xx')}
        \big].
\end{align*}
The PSD kernel can be rewritten as
\begin{align}
    \label{equation.psd-kernel}
    k(\xx, \xx') = \int_{\mathbb{R}^d \times \mathbb{R}^d} \mathcal{E}_{\oo, \oo'}(\xx, \xx') \mu (d \oo, d \oo').
\end{align}
where $s(\oo, \oo')$ is the spectral density surface.
Similar spectral representation forms for non-stationary kernels are also used in \cite{samo2015generalized,remes2017non}.
When $\oo = \oo'$, non-stationary kernel in (\ref{equation.psd-kernel}) degrades into stationary kernel as in Bochner's theorem (\ref{equation.stationary}) and the probability density function becomes univariate $s(\oo)$.

In the stationary cases, random Fourier features are used to approximate stationary kernels \cite{rahimi2007random}.
Similarly, in the non-stationary cases, we can approximate (\ref{equation.psd-kernel}) with Monte Carlo random sampling
\begin{align*}
              & k(\xx, \xx') = \int_{\mathbb{R}^d \times \mathbb{R}^d} \mathcal{E}_{\oo, \oo'}(\xx, \xx') \mu (d \oo, d \oo') \\
    = ~       & \mathbb{E}_{\oo, \oo' \sim s} ~~ \big[\mathcal{E}_{\oo, \oo'}(\xx, \xx')\big]                                     \\
    = ~       & \mathbb{E}_{\oo, \oo' \sim s} ~~ \frac{1}{4}\big[\cos(\oo^T\xx - \oo'^T\xx') + \cos(\oo'^T\xx - \oo^T\xx')        \\
              & + ~\cos(\oo^T\xx - \oo^T\xx') + \cos(\oo'^T\xx - \oo'^T\xx')\big]                                                  \\
    \approx ~ & \frac{1}{4D} \sum_{i=1}^D \Big[\cos(\oo_i^T\xx - \oo_i'^T\xx') + \cos(\oo_i'^T\xx - \oo_i^T\xx')                  \\
              & + ~\cos(\oo_i^T\xx - \oo_i^T\xx') + \cos(\oo_i'^T\xx - \oo_i'^T\xx')
        \Big]                                                                                                                     \\
    = ~       & \langle \psi(\xx), \psi(\xx')\rangle
\end{align*}
where ${(\oo_i, \oo_i')}_{i=1}^D \overset{\text{i.i.d.}}{\sim} s(\oo, \oo')$, $D$ is the number of features and random Fourier feature mapping of spectral kernel is
\begin{align*}
    \psi(\xx) = \frac{1}{\sqrt{4D}}
    \begin{bmatrix}
        \cos(\OO^T \xx) + \cos(\OO'^T \xx) \\
        \sin(\OO^T \xx) + \sin(\OO'^T \xx)
    \end{bmatrix},
\end{align*}
where features mapping are $\mathbb{R}^d \to \mathbb{R}^{2D}$.
To alleviate computational costs, we use the following mapping $\mathbb{R}^d \to \mathbb{R}^{D}$
\begin{align}
    \label{equation.non-startionary-rff}
    \phi(\xx) = \frac{1}{\sqrt{2D}}
    \begin{bmatrix}
        \cos(\OO^T \xx + {\boldsymbol b}) + \cos(\OO'^T \xx + {\boldsymbol b}')
    \end{bmatrix},
\end{align}
where frequency matrices $\OO, \OO' \in \mathbb{R}^{d \times D}$ are integrated with ${(\oo_i, \oo_i')}_{i=1}^D$ that $\OO = [\oo_1, \cdots, \oo_D]$ and $\OO' = [\oo'_1, \cdots, \oo'_D]$.
The phase vectors ${\boldsymbol b}, {\boldsymbol b}'$ are drawn uniformly from $[0, 2\pi]^D$.
In fact, the spectral kernels induced by $\psi(\xx)$ and $\phi(\xx)$ are equivalent in the expectation manner.


\begin{remark}
    Equation (\ref{equation.non-startionary-rff}) provides explicit feature mappings to approximate non-stationary kernels, where suitable frequency matrices $\OO, \OO'$ are the key to obtain a favorable performance.
    Standard random Fourier features generate frequency matrices via samples from the \textbf{assigned} spectral density $s(\oo, \oo')$, e.g frequency matrices correspond to random Gaussian matrix for Gaussian kernels.
    In the proposed algorithm \texttt{ASKL}, the frequency matrices $\OO, \OO'$ are jointly \textbf{optimized} together with model weights $\WW$ during training.
\end{remark}

\subsection{Update Trainable Matrices}
As shown in the previous section, non-stationary spectral kernels are input-dependent.
Further, we propose a general learning framework, namely Automated Spectral Kernel Learning (\texttt{ASKL}), that \textbf{optimizes} frequency matrices $\OO, \OO'$ for non-stationary kernels according to the objective via backpropagation.
Therefore, the learned feature mappings are both \textit{input-dependent} and \textit{output-dependent}.

The learning frame \texttt{ASKL} can be solved by first-order gradient descent methods, such as stochastic gradient descent (SGD) and its variants Adadelta \cite{zeiler2012adadelta} and Adam \cite{kingma2014adam}.
Using the backpropagation, we derive how to update estimator weights $\WW$ and frequency matrices $\OO, \OO'$ in the following analyses.
More generally, we consider mini-batch gradient descent where we use $m$ examples in each iteration.
Specifically, full gradient descent is the special case $m=n$ where all examples are used and stochastic gradient descent (SGD) corresponds to the special case $m=1$ where only one example is used.

\subsubsection{Update $\WW$ in Proximal Gradient Approach}
To minimize the objective (\ref{equation.primal-objective}), we use first-order gradient descent algorithms to update $\WW$ in the direction of negative gradient.
The updates of gradient of $\WW$ depend on empirical loss and trace norm in (\ref{equation.primal-objective}), but trace norm is nondifferentiable on many points for each dimension (unlike hinge loss and Relu are nondifferentiable only on one point), thus the derivative/subgradient of the trace norm cannot be applied to standard gradient approaches.
We employ singular value thresholding (SVT) to solve the minimization of trace norm with proximal gradient descent \cite{cai2010singular}.

We simply the updates of $\WW$ in two steps and put detailed deduction process in the Appendix section:
\begin{itemize}
    \item[1)] Update $\WW$ with SGD on the empirical loss
    \begin{align}
        \label{gradient.W}
        {\boldsymbol Q} = \WW^{t}-\eta\nabla g(\WW^{t}),
    \end{align}
    where $\eta$ is the learning rate and the gradient of empirical loss on $m$-batch examples is
    \begin{equation*}
        \begin{aligned}
             \nabla g(\WW) &= ~\frac{1}{m}\sum_{i=1}^m \frac{\partial \ell(f(\xx_i), \yy)}{\partial \WW}                                                        \\
             & =~\frac{1}{m}\sum_{i=1}^m \phi(\xx_i) \cdot \left[\frac{\partial \ell(f(\xx_i), \yy)}{\partial f(\xx_i)}\right]^T \in \mathbb{R}^{D \times K},
        \end{aligned}
    \end{equation*}
    and ${\boldsymbol Q}$ is an intermediate matrix and $m$ examples are used.
    \item[2)] Update $\WW$ with SVT on the trace norm
        \begin{align}
            \label{equation.svt_result}
            \WW^{t+1} = {\boldsymbol U}\text{diag}\big(\left\{\sigma_j - \lambda_1\eta\right\}_+\big){\boldsymbol V}^T,
        \end{align}
        where ${\boldsymbol Q}={\boldsymbol U}{\boldsymbol \Sigma}{\boldsymbol V}^T$ is the singular values decomposition, ${\boldsymbol \Sigma}$ is the diagonal ${\text{diag}(\{\sigma_j\}_{1 \leq i \leq r})}$ and $r$ is the rank of ${\boldsymbol Q}$.
\end{itemize}
\subsubsection{Update $\OO, \OO'$}
Using the chain rule for computing the derivative, the derivative of objective only depends on empirical loss $\ell(f(\xx_i), \yy)$ in terms of $\OO, \OO'$.
Both empirical risk $\ell(f(\xx_i), \yy)$ and squared Frobenius norm $\|\phi(\boldsymbol{X})\|_F^2$ are differentiable with respect to $\OO, \OO'$.
We derive the gradient of the minimization objective \eqref{equation.primal-objective} w.r.t. $\OO$ as an example.
\begin{equation}
    \label{gradient.Omega}
    \begin{aligned}
        \OO^{t+1} &= \OO^{t}-\eta \nabla \widehat{\mathcal{E}}(\OO^{t})
    \end{aligned}
\end{equation}
where the gradients w.r.t $\OO$ are
\begin{align*}
    &\nabla \widehat{\mathcal{E}}(\OO) = 
        \frac{1}{m}\sum_{i=1}^m \frac{\partial\ell(f(\xx_i), \yy)}{\partial \OO}
        + \lambda_2 \frac{\partial\|\phi(\boldsymbol{X})\|_F^2}{\partial \OO}, \\
        & \frac{\partial\ell(f(\xx_i), \yy)}{\partial \OO} = \xx_i \cdot \left[ \boldsymbol{D} \cdot \WW \cdot \frac{\partial\ell(f(\xx_i), \yy)}{\partial f(\xx_i)} \right]^T, \\
        & \frac{\partial\|\phi(\boldsymbol{X})\|_F^2}{\partial \OO}
       = \frac{1}{m}\sum_{i=1}^m 2\xx_i \cdot \phi(\xx_i)^T \cdot \boldsymbol{D}
\end{align*}
and $\boldsymbol{D}$ is a diagonal matrix in $D \times D$ size filled with a vector
\begin{align*}
    \boldsymbol{D} = \text{diag}\left\{\frac{-1/\sqrt{2D}}{ \sin(\OO^T\xx_i + b)}\right\}_{D \times D}.
\end{align*}

\subsubsection{Specific Loss Functions} For gradients in (\ref{gradient.W}) and (\ref{gradient.Omega}), only gradients w.r.t. the loss function $\frac{\partial\ell(f(\xx_i), \yy)}{\partial f(\xx_i)}$ are uncertain. 
\begin{itemize}
    \item {Hinge Loss for Classification Problems.}\\
          Let the label $\yy = [0, \cdots, 0, 1, 0, \cdots, 0]^T$ only one element (its category) is not zero.
          The hinge loss is defined as
          $\ell(f(\xx_i), \yy) = |1-(\yy^Tf(\xx_i)-\max_{\yy'\not=\yy} \yy'^Tf(\xx_i))|_+$.
          The sub-gradient of hinge loss w.r.t the estimator is
          \begin{equation*}
              \frac{\partial\ell(f(\xx_i), \yy)}{\partial f(\xx_i)} =
              \left\{
              \begin{aligned}
                   & \mathbf{0}, \quad \yy^Tf(\xx_i)-\max_{\yy'\not=\yy} \yy'^Tf(\xx_i) \geq 1, \\
                   & \yy' - \yy, \quad       \text{else}.
              \end{aligned}
              \right.
          \end{equation*}
    \item {Squared Loss for Regression Problems.}\\
          Let $\yy$ be the $K$-size vector-valued label where $K > 1$ for multi-label regression and $K = 1$ for univariate regression.
          The Squared loss function is $\ell(f(\xx_i), \yy) = \|f(\xx_i) - \yy\|_2^2.$
          Then, the gradient of squared loss is
          \begin{equation*}
              \frac{\partial\ell(f(\xx_i), \yy)}{\partial f(\xx_i)} = 2(f(\xx_i) - \yy).
          \end{equation*}
\end{itemize}

\section{Theoretical Guarantee}
In this section, we study the generalization performance for the proposed algorithm \texttt{ASKL}.
A data-dependent excess risk bound is derived, and then we explore how the input-dependent and output-dependent feature mappings affect the statistical performance and how to improve the algorithm by utilizing additional regularization terms.

\begin{definition}
    The empirical Rademacher complexity of hypothesis space $\mathcal{H}$ is
    \begin{align*}
         & \widehat{\mathcal{R}}(\mathcal{H})
        = \frac{1}{n} ~ \mathbb{E}_\epsilon \left[\sup_{f \in \mathcal{H}} \sum_{i=1}^{n} \sum_{k=1}^K \epsilon_{ik} f_k(\xx_i)\right],
    \end{align*}
    where $f_k(\xx_i)$ is the $k$-th value of the estimator $f(\xx_i)$ with $K$ outputs and $\epsilon_{ik}$s are $n \times K$ independent Rademacher variables with equal probabilities $\mathbb{P}(\epsilon_{ik}=+1)=\mathbb{P}(\epsilon_{ik}=-1)=1/2$.
    Its deterministic estimate is $\mathcal{R}(\mathcal{H})=\mathbb{E} ~ \widehat{\mathcal{R}}(\mathcal{H})$.
\end{definition}

\begin{theorem}[Excess Risk Bound]
    \label{thm.excess_risk_bounds}
    Assume that $B = \sup_{f \in \mathcal{H}} \|\WW\|_* < \infty$ and assume the loss function $\ell$ is $L$-Lipschitz for $\mathbb{R}^K$ equaipped with the $2$-norm, with probability at least $1-\delta$, the following excess risk bound holds
    \begin{align}
        \label{inequation.grc-bound-result-excess-risk}
        \mathcal{E}(\widehat{f}_n) - \mathcal{E}(f^*) \leq 4 \sqrt{2} L \widehat{\mathcal{R}}(\mathcal{H}) + \mathcal{O}\Big(\sqrt{\frac{\log 1/\delta}{n}}\Big),
    \end{align}
    where $f^* \in \mathcal{H}$ is the most accurate estimator in the hypothesis space, $\widehat{f}_n$ is the empirical estimator and
    \begin{equation}
        \label{inequation.grc-bound-result-rff}
        \begin{aligned}
             & \widehat{\mathcal{R}}(\mathcal{H})
            \leq ~ \frac{B}{n} \sqrt{K \sum_{i=1}^n \langle \phi(\xx_i), \phi(\xx_i) \rangle}                                                 \\
             & = ~ \frac{B}{n} \sqrt{ \frac{K}{D} \sum_{i=1}^n \sum_{j=1}^D \frac{1}{2}\Big[\cos\big((\oo_j - \oo_j')^T \xx_i\big) + 1\Big]}.
        \end{aligned}
    \end{equation}
\end{theorem}
The proof is given in the Appendix.
The error bounds depend on Rademacher complexity term.
Due to $\cos\big((\oo_j - \oo_j')^T \xx_i\big) \leq 1$, Rademacher complexity in (\ref{inequation.grc-bound-result-rff}) is naturally bounded by
$
    \widehat{\mathcal{R}}(\mathcal{H}) \leq B \sqrt{ {K}/{n}},
$
thus the convergence rate is
\begin{align}
    \label{inequation.convergence-rate}
    \mathcal{E}(\widehat{f}_n) - \mathcal{E}(f^*) \leq \mathcal{O}\left(B\sqrt{\frac{K}{n}}\right).
\end{align}
Based on the theoretical findings, we make some technical comments to understand how the factors, including feature mappings and regularization terms, affect the generalization performance and suggest ways to refine the algorithm:
\begin{itemize}
    \item \textbf{Influence of Non-Stationary Kernels.}
          As mentioned in the above section, non-stationary kernels depend on inputs themselves instead of the distance between inputs.
          We explore the impact of non-stationary kernels based on Rademacher complexity in (\ref{inequation.grc-bound-result-rff}), which depends on the trace $\sum_{i=1}^n k(\xx_i, \xx_i)$.
          For stationary kernels (shift-invariant kernels), the spectral representation holds the diagonals as $k(\xx_i, \xx_i) = \cos(\oo^T(\xx_i -\xx_i)) = 1$, thus the trace of kernel matrix $\sum_{i=1}^n k(\xx_i, \xx_i) = n,$ which corresponds to the worst cases.
          While for non-stationary kernels, $k(\xx_i, \xx_i) = \cos((\oo - \oo')^T\xx_i) \in [-1, 1].$
          For most instances $\xx_i$, the diagonals are $k(\xx_i, \xx_i) < 1$, so the trace $\sum_{i=1}^n k(\xx_i, \xx_i) \ll n$.
          Therefore, error bounds of non-stationary spectral kernels are much tighter than bounds of stationary kernels, where non-stationary kernels achieve the better generalization performance.
    \item \textbf{Influence of Spectral Learning.}
          If frequency matrices $\OO, \OO'$ are just assigned according to specific spectral density $s(\oo, \oo')$, the non-stationary kernels are input-dependent but output-independent.
          By dynamically optimizing frequency matrices $\OO, \OO'$ towards the objective, we acquire more powerful feature representations.
          Then, the spectral measure $s(\oo, \oo')$ of optimal kernels can be estimated from optimized frequency matrices $\OO, \OO'$.
          The learned feature mappings are dependent on both inputs and outputs, offering better feature representations.
    \item \textbf{Using the Trace Norm $\|\WW\|_*$ as a Regularizer.}
          The convergence rate $B = \sup_{f \in \mathcal{H}} \|\WW\|_* < \infty$ is also dependent on a constant $B$, that is the supremum of  trace norm $\|\WW\|_*$ in terms of the specific hypothesis space.
          As a  result, the minimization of trace norm $\|\WW\|_*$ is useful to reduce $B$ and obtain better error bounds.
          Based on Rademacher complexity theory, the use of trace norm $\|\WW\|_*$ instead of squared Frobenius norm $\|\WW\|_F^2$ as regularization terms was also explored for linear estimators in \cite{xu2016local,li2019multi}.
    \item \textbf{Using Squared Frobenius Norm $\|\phi(\boldsymbol{X})\|_F^2$ as a Regularizer.}
          From (\ref{inequation.grc-bound-result-excess-risk}), Rademacher complexity is bounded by the trace of the kernel and it can be written as
          \begin{align*}
              \sum_{i=1}^n \langle \phi(\xx_i), \phi(\xx_i) \rangle = \sum_{i=1}^n \|\phi(\xx_i)\|_2^2 = \|\phi(\boldsymbol{X})\|_F^2.
          \end{align*}
          In standard theoretical learning for Rademacher complexity, the kernel trace cannot be used to improve learning algorithm because feature mappings are constants for specific inputs.
          For example, all diagonals are $k(\xx_i, \xx_i) = 1$ for shift-variant kernels thus the trace is the number of examples $n$.
          Note that, instead of assigned kernel parameters, our algorithm automatically learns optimal spectral density for kernels, thus the trace of the kernel is no longer a constant.
          Meanwhile, we use it as two regularization terms to improve the performance (avoid overfitting).
\end{itemize}


\section{Experiments}
In this section, compared with other algorithms, we evaluate the empirical behavior of our proposed algorithm \texttt{ASKL} on several benchmark datasets to demonstrate the effects of factors used in our algorithm, including the non-stationary spectral kernel, updating spectral density with backpropagation and additional regularization terms.
\subsection{Experimental Setup}
Based on random Fourier features, both our algorithm \texttt{ASKL} and compared methods apply nonlinear feature mapping into a fixed $D$-dimensional feature space where we set $D=2000$.
We apply Gaussian kernels as basic kernels because Gaussian kernels succeed in many types of data by mapping inputs into infinite-dimensional space, of which frequency matrices $\OO, \OO'$ are i.i.d. drawn from Gaussian distributions $\mathcal{N}(0, \sigma^2)$.
The generalization ability of algorithms are highly dependent on different parameters on $\lambda_1, \lambda_2$ and Gaussian kernel parameter $\sigma$ for spectral kernels.
For fair comparisons, we tune those parameters to achieve optimal empirical performance for all algorithms on all dataset, by using $5$-folds cross-validation and grid search over parameters candidate sets.
Regularization parameters are selected in $\lambda_1, \lambda_2 \in \{10^{-10}, 10^{-9}, \cdots, 10^{-1}\}$ and Gaussian kernel parameter $\sigma$ is selected from candidate set $\sigma \in \{2^{-10}, \cdots, 2^{10}\}$.
Accuracy and mean squared error (MSE) are used to evaluate performance for classification and regression, respectively.
We implement all algorithms based on Pytorch and use Adam as optimizer with $32$ examples in a mini-batch to solve the minimization problem.

\subsubsection{Compared Algorithms}
\begin{table}[t]
    \centering
    \resizebox{1\columnwidth}{!}{
    \setlength\extrarowheight{3pt}
    \begin{tabular}{llll}
        \toprule
        Methods     & Kernel         & Density & Regularizer \\ \hline
        SK            & Stationary     & Assigned & $\|\WW\|_F^2$        \\
        NSK           & Non-stationary & Assigned & $\|\WW\|_F^2$       \\ \hline
        SKL           & Stationary     & Learned  & $\|\WW\|_F^2$        \\
        NSKL          & Non-stationary & Learned  & $\|\WW\|_F^2$        \\ \hline
        \texttt{ASKL} & Non-stationary & Learned  & $\|\WW\|_*, \|\phi(\boldsymbol{X})\|_F^2$   \\
        \bottomrule
    \end{tabular}
    }
    \caption{Compared algorithms.}
    \label{tab.compared-methods}
\end{table}

To assess the effectiveness of factors used in our algorithm, we compare the proposed algorithm with several relevant algorithms.
As shown in Table \ref{tab.compared-methods}, compared methods are special cases of \texttt{ASKL}:\\
(1) \textbf{SK} \cite{rahimi2007random}: known as random Fourier features for the stationary spectral kernel.
This approach directly assigned the spectral density for shift-invariant kernels and uses the regularizer $\|\WW\|_F^2$ on model weights .\\
(2) \textbf{NSK} \cite{samo2015generalized}: Similar to \textbf{SK} but it uses spectral representation for non-stationary kernel which was introduced in (\ref{equation.non-startionary-rff}) with assigned frequency matrices.\\
(3) \textbf{SKL} \cite{huang2014kernel}: Random Fourier features for stationary kernels and squared Frobenius norm as regularization term with updating spectral density during training.\\
(4) \textbf{NSKL}: A special case of \texttt{ASKL} with non-stationary spectral, learned spectral density.
But it uses squared Frobenius norm on model weights $\|\WW\|_F^2$ as regularization.

\subsubsection{Datasets}
We evaluate the performance of the proposed learning framework \texttt{ASKL} and compared algorithms based on several publicly available datasets, including both classification and regression tasks.
Especially, we standardize outputs for regression tasks to $[0, 100]$ for better illustration.
To obtain stable results, we run methods on each dataset 30 times with randomly partition such that 80\% data for training and $20\%$ data for testing.
Further, those multiple test errors allow the estimation of the statistical significance of difference among methods.
To explore the influence of factors upon convergence, we evaluate both test accuracy and objective on MNIST dataset \cite{lecun1998gradient}.

\subsection{Empirical Results}
\begin{table*}[t]
    \centering
    \setlength\extrarowheight{1.1pt}
    \begin{tabular}{@{\extracolsep{0.6cm}}cl|cc|cc|c}
        \toprule
         &           & SK             & NSK            & SKL                        & NSKL                       & \texttt{ASKL}             \\ \hline
        \multirow{8}*{Accuracy($\uparrow$)}
         & segment   & 89.93$\pm$2.12 & 90.15$\pm$2.08 & 94.58$\pm$1.86             & 94.37$\pm$0.81             & \textbf{95.02$\pm$1.54}   \\
         & satimage  & 74.54$\pm$1.35 & 75.15$\pm$1.38 & 83.61$\pm$1.08             & 83.74$\pm$1.34             & \textbf{85.32$\pm$1.45}   \\
         & USPS      & 93.19$\pm$2.84 & 93.81$\pm$2.13 & 95.13$\pm$0.91             & 95.27$\pm$1.65             & \textbf{97.76$\pm$1.14}   \\
         & pendigits & 96.93$\pm$1.53 & 97.39$\pm$1.41 & 98.19$\pm$2.30             & 98.28$\pm$1.68             & \textbf{99.06$\pm$1.26}   \\
         & letter    & 76.50$\pm$1.21 & 78.21$\pm$1.56 & 93.60$\pm$1.14             & 94.66$\pm$2.21             & \textbf{95.70$\pm$1.74}   \\
         & porker    & 49.80$\pm$2.11 & 51.85$\pm$0.97 & 54.27$\pm$2.72             & \underline{54.69$\pm$1.68} & \textbf{54.85$\pm$1.28}   \\
         & shuttle   & 98.17$\pm$2.81 & 98.21$\pm$1.46 & \underline{98.87$\pm$1.42} & 98.74$\pm$1.07             & \textbf{98.98$\pm$0.94}   \\
         & MNIST     & 96.03$\pm$2.21 & 96.45$\pm$2.16 & 96.67$\pm$1.61             & 98.03$\pm$1.16             & \textbf{98.26$\pm$1.78}   \\
        \hline
        \multirow{4}*{RMSE($\downarrow$)}
         & abalone   & 10.09$\pm$0.42 & 9.71$\pm$0.28  & 8.35$\pm$0.28              & \textbf{7.85$\pm$0.42}     & \underline{7.88$\pm$0.16} \\
         & space\_ga & 11.86$\pm$0.26 & 11.58$\pm$0.42 & 11.40$\pm$0.18             & 11.39$\pm$0.46             & \textbf{11.34$\pm$0.27}   \\
         & cpusmall  & 2.77$\pm$0.71  & 2.84$\pm$0.38  & 2.56$\pm$0.72              & 2.57$\pm$0.63              & \textbf{2.42$\pm$0.48}    \\
         & cadata    & 50.31$\pm$0.92 & 51.47$\pm$0.32 & 47.67$\pm$0.33             & 47.71$\pm$0.30             & \textbf{46.34$\pm$0.23}   \\
        \bottomrule
    \end{tabular}
    \caption{
        \normalsize  Classification accuracy (\%) for classification datasets and RMSE for regression datasets. ($\uparrow$) means the lager the better while ($\downarrow$) indicates the smaller the better. We bold the numbers of the best method and underline the numbers of the other methods which are not significantly worse than the best one. }
    \label{tabel.accuracy}
\end{table*}

Empirical results of all algorithms are shown in Table \ref{tabel.accuracy}, where accuracy is used for classification tasks and root mean squared error (RMSE) is used for regression tasks.
We bold results which have the best performance on each dataset, but also mark sub-optimal results with underlines which have a significant difference with the best ones, by using pairwise $t$-test on results of 30 times repeating data split and training.

The results in Table \ref{tabel.accuracy} show:
(1) The proposed algorithm \texttt{ASKL} outperforms compared algorithms on almost all dataset that coincides with our theoretical results.
(2) The use of non-stationary kernels brings notable performance improvement but approaches based on stationary kernels still perform well on easy tasks, e.g. \textit{shuttle} and \textit{cpusmall}.
(3) Due to the difference between assigned spectral density and learned frequency matrices, there are significant performance gaps between those two groups \{SK, NSK\} and \{SKL, NSKL\}, especially on complicated datasets \textit{satimage} and \textit{letter}.
It confirms that learning feature mappings in both input-dependent and output-dependent ways leads to better generalization performance.
(4) The proposed algorithm \texttt{ASKL} usually provides better results than NSKL.
That shows the effectiveness of regularization terms $\|\WW\|_*$ and $\|\phi(\boldsymbol{X})\|_F^2$.
The experimental results demonstrate the excellent performance and stability of \texttt{ASKL}, corroborating the theoretical analysis and excellent performance of \texttt{ASKL}.

\begin{figure}[t]
    \begin{center}
        \includegraphics[width=.83\columnwidth]{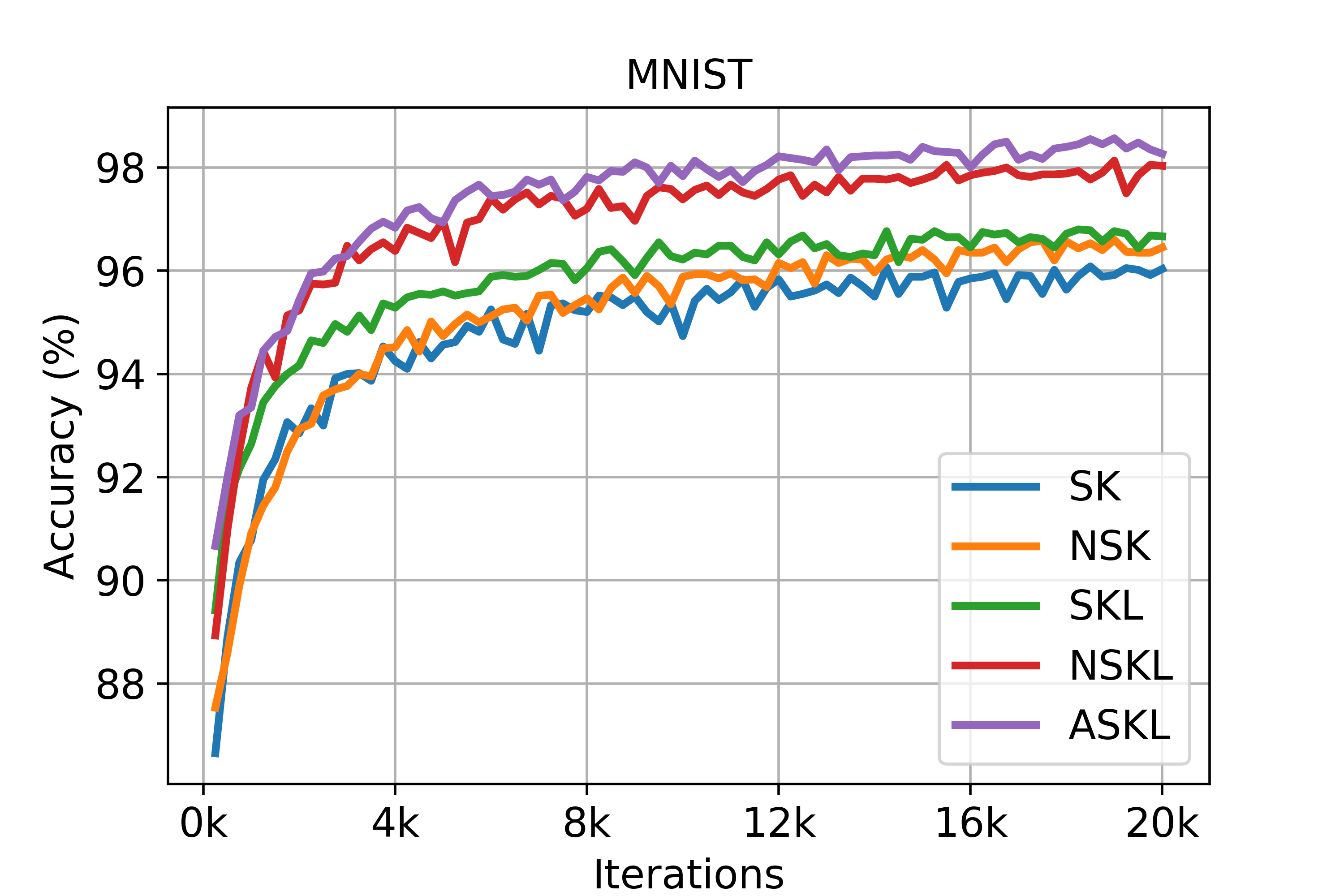}
    \end{center}
    \caption{Accuracy curves on MNIST}
    \label{fig.accuracy}
\end{figure}

\begin{figure}[t]
    \begin{center}
        \includegraphics[width=.83\columnwidth]{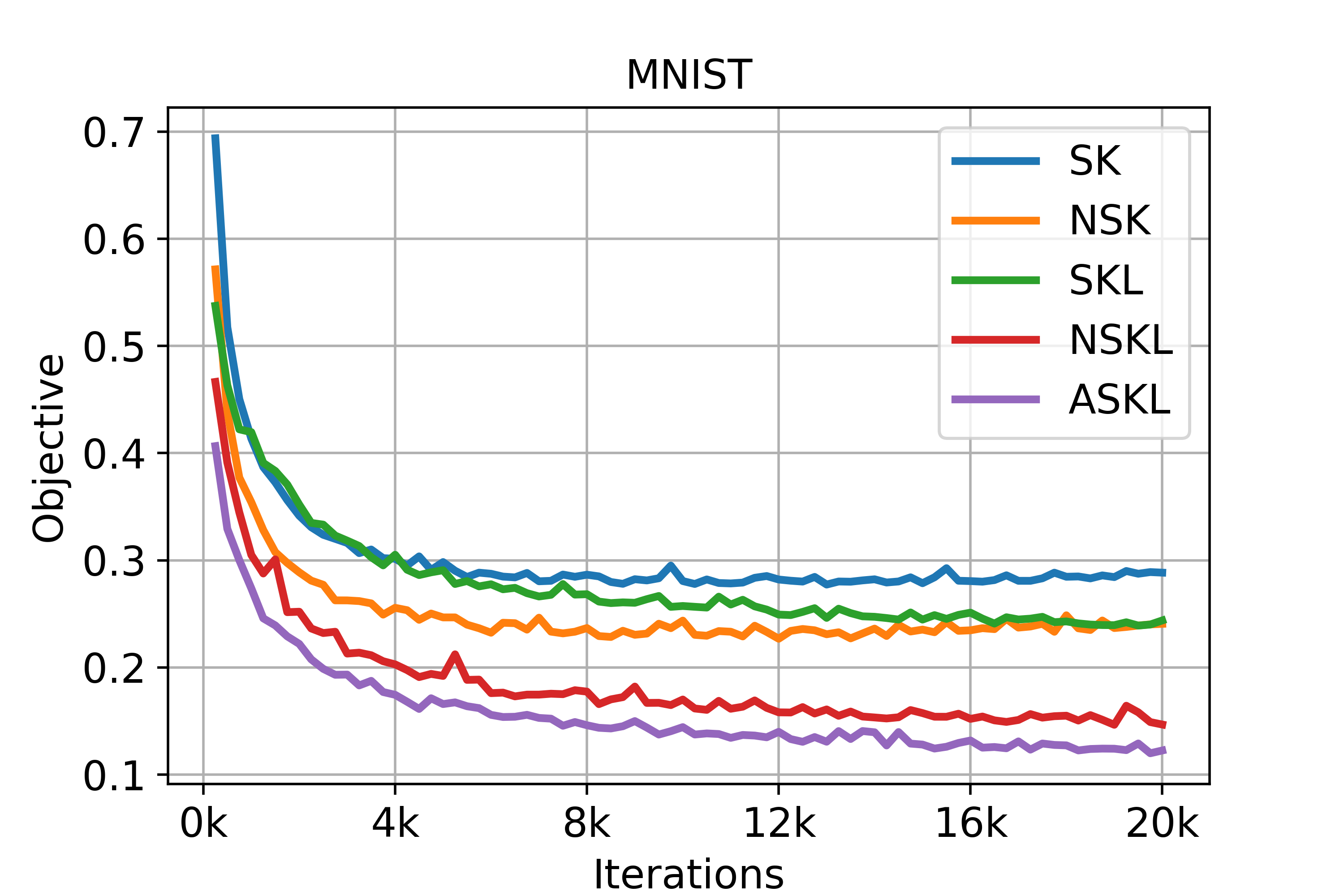}
    \end{center}
    \caption{Objective curves on MNIST}
    \label{fig.objective}
\end{figure}

During iterations, test accuracy and the objective were recorded for every 200 iterations with batch size 32.
Evaluation results in Figure \ref{fig.accuracy} and Figure \ref{fig.objective} show that \texttt{ASKL} outperforms other algorithms significantly.
Accuracy curves and objective curves are correlated as the smaller objective corresponds to the higher accuracy.
Figure \ref{fig.accuracy} and Figure \ref{fig.objective} empirically illustrate that \texttt{ASKL} achieves lower error bounds than with fast learning rates.

\section{Conclusion and Discussion}
In this paper, we propose automatically kernel learning framework, which jointly learns spectral density and the estimator.
Both theoretical analysis and experiments illustrate that the framework obtains significant improvements in the generalization performance, owing to the use of three factors: non-stationary spectral kernel, backpropagation and two regularization terms.
The use of non-stationary spectral kernels makes feature mappings \textit{input-dependent}, while updating frequency matrices w.r.t the objective via backpropagation that guarantees feature mappings are \textit{output-dependent}, obtaining more powerful representation abilities.
Further, we derive Rademacher complexity bounds for the algorithm.
To achieve sharper bounds, we minimize two regularization terms together with ERM.

\textbf{Connection with Deep Neural Networks.}
The proposed learning framework is also a neural network with a single hidden layer and the cosine activation, thus the theoretical findings are also applied to this kind of neural networks.
Those results can be extended to deep neural networks (DNN) by stacking $\phi$ in the hierarchical structure.
For example $l$-hidden layers are used that the spectral kernel is $k(\xx, \xx') = \langle \phi^l(\phi^{l-1}(\cdots \phi^1(\xx))), \phi^l(\phi^{l-1}(\cdots \phi^1(\xx'))) \rangle$.
The outputs of other activations such as sigmoid and Relu can also be seen as random feature mapping $\varphi(\xx)$, corresponding kernel function is $k(\xx, \xx') = \langle \varphi(\xx), \varphi(\xx')\rangle$.
It is a possible way to understand deep neural networks in kernels view based on Rademacher complexity theory.


\section{Appendix}
Firstly, we introduce common notations used in Rademacher complexity theory.
The space of loss function associated with $\mathcal{H}$ is denoted by
\begin{align}
    \label{eq.loss-space}
    \mathcal{L}=\left\{\ell(f(\xx), \yy) ~ \big | ~ f\in\mathcal{H}\right\}.
\end{align}
\begin{definition}[Rademacher complexity on loss space]
    \label{def.rc-loss}
    Assume $\mathcal{L}$ is a space of loss functions as defined in Equation \eqref{eq.loss-space}.
    Then the empirical Rademacher complexity of $\mathcal{L}$ is:
    \begin{align*}
         & \widehat{\mathcal{R}}(\mathcal{L})=
        \frac{1}{n} ~ \mathbb{E}_{\epsilon}
        \left[\sup_{\ell\in\mathcal{L}}
            \sum_{i=1}^n\epsilon_i \ell(f(\xx_i), \yy_i)
            \right],
    \end{align*}
    where $\epsilon_i$s are independent Rademacher random variables uniformly distributed over $\{\pm1\}$.
    Its deterministic counterpart is $\mathcal{R}(\mathcal{L})=\mathbb{E} ~ \widehat{\mathcal{R}}(\mathcal{L})$.
\end{definition}

\begin{lemma}[Lemma 5 of \cite{cortes2016structured}]
    \label{lem.contraction}
    Let the loss function $\ell$ be $L$-Lipschitz for $\mathbb{R}^K$ equaipped with the $2$-norm,
    \begin{align*}
        |\ell(f(\xx), \yy) - \ell(f(\xx'), \yy')| \leq L\|f(\xx) - f(\xx')\|_2.
    \end{align*}
    Then, the following contraction inequation exists
    \begin{align*}
        \mathcal{R} (\mathcal{L}) \leq \sqrt{2} L \mathcal{R}(\mathcal{H}).
    \end{align*}
\end{lemma}

\subsection{Proof of Theorem \ref{thm.excess_risk_bounds}}
\begin{proof}
    A standard fact is the derivation of expected loss and empirical means can be controlled by the Rademacher averages over loss space $\mathcal{L}$ (Lemma A.5 of \cite{bartlett2005local})
    \begin{align}
        \label{inequation.grc-bound-excess-risk}
        \mathcal{E}(\widehat{f}_n) - \mathcal{E}(f^*) \leq 2 ~ \Big[\sup_{f \in \mathcal{H}} ~ \widehat{\mathcal{E}}(f) - \mathcal{E}(f)\Big] \leq 4 \mathcal{R}(\mathcal{L}).
    \end{align}
    Combining (\ref{inequation.grc-bound-excess-risk}) with the contraction in Lemma \ref{lem.contraction} and connection between $\widehat{\mathcal{R}}(\mathcal{H})$ and $\mathcal{R}(\mathcal{H})$ (Lemma A.4 of \cite{bartlett2005local}), there holds with high probability at least $1-\delta$
    \begin{align}
        \label{inequation.grc-bound-empirical-grc}
        \mathcal{E}(\widehat{f}_n) - \mathcal{E}(f^*) \leq 4 \sqrt{2} L \widehat{\mathcal{R}}(\mathcal{H}) + \mathcal{O}\Big(\sqrt{\frac{\log 1/\delta}{n}}\Big).
    \end{align}
    Then, we estimate empirical Rademacher complexity $\widehat{\mathcal{R}}(\mathcal{H})$
    \begin{equation}
        \label{equation.grc-bound-grc-estimate}
        \begin{aligned}
            \widehat{\mathcal{R}}(\mathcal{H}) & = \frac{1}{n} ~ \mathbb{E}_{\boldsymbol \epsilon} \left[\sup_{f \in \mathcal{H}} \sum_{i=1}^n \sum_{k=1}^K \epsilon_{ik} f_k(\xx_i)\right]             \\
                                               & = \frac{1}{n} ~ \mathbb{E}_{\boldsymbol \epsilon} \left[\sup_{f \in \mathcal{H}} \langle \WW, \boldsymbol{\Phi}_{\boldsymbol \epsilon} \rangle \right]
        \end{aligned}
    \end{equation}
    where $\WW, \boldsymbol{\Phi}_{\boldsymbol \epsilon} \in \mathbb{R}^{D \times K}$ and $\langle \WW, \boldsymbol{\Phi}_{\boldsymbol \epsilon} \rangle = \text{Tr}(\WW^T\boldsymbol{\Phi}_{\boldsymbol \epsilon})$, we define the matrix $\boldsymbol{\Phi}_{\boldsymbol \epsilon}$ as follows:
    \begin{align*}
        \boldsymbol{\Phi}_{\boldsymbol \epsilon} := \left[\sum_{i = 1}^n \epsilon_{i1}\phi(\xx_i),
            \sum_{i = 1}^n \epsilon_{i2}\phi(\xx_i),
            \cdots, \sum_{i = 1}^n \epsilon_{iK}\phi(\xx_i)\right].
    \end{align*}
    Applying H\"older's inequality and $\|\WW\|_*$ bounded by a constant $B$ to (\ref{equation.grc-bound-grc-estimate}), we can obtain
    \begin{equation}
        \label{inequation.grc-bound-holder}
        \begin{aligned}
             & \widehat{\mathcal{R}}(\mathcal{H})
            = ~   \frac{1}{n} ~ \mathbb{E}_{\boldsymbol \epsilon} \left[\sup_{f \in \mathcal{H}} \langle \WW, \boldsymbol{\Phi}_{\boldsymbol \epsilon} \rangle \right] \\
            &\leq ~ \frac{1}{n} ~ \mathbb{E}_{\boldsymbol \epsilon} \left[\sup_{f \in \mathcal{H}} \|\WW\|_* \|\boldsymbol{\Phi}_{\boldsymbol \epsilon}\|_F \right]        
             \leq \frac{B}{n} ~ \mathbb{E}_{\boldsymbol \epsilon} \left[\|\boldsymbol{\Phi}_{\boldsymbol \epsilon}\|_F \right] \\
            &\leq ~ \frac{B}{n} ~ \mathbb{E}_{\boldsymbol \epsilon} \left[\sqrt{\|\boldsymbol{\Phi}_{\boldsymbol \epsilon}\|_F^2} \right]
            \leq \frac{B}{n} ~ \sqrt{\mathbb{E}_{\boldsymbol \epsilon} ~ \|\boldsymbol{\Phi}_{\boldsymbol \epsilon}\|_F^2}.
        \end{aligned}
    \end{equation}
    Then, we bound $\mathbb{E}_{\boldsymbol \epsilon} ~ \|\boldsymbol{\Phi}_{\boldsymbol \epsilon}\|_F^2$ as follows
    \begin{equation}
        \label{inequation.grc-bound-estimate-Phi}
        \begin{aligned}
             & \mathbb{E}_{\boldsymbol \epsilon} ~ \|\boldsymbol{\Phi}_{\boldsymbol \epsilon}\|_F^2
            \leq ~  \mathbb{E}_{\boldsymbol \epsilon} ~ \sum_{k=1}^K \Big\|\sum_{i=1}^n \epsilon_{ik}\phi(\xx_i)\Big\|_2^2                                          \\
             & \leq ~  \sum_{k=1}^K \mathbb{E}_{\boldsymbol \epsilon} ~ \Big\|\sum_{i=1}^n \epsilon_{ik}\phi(\xx_i)\Big\|_2^2                                       \\
             & \leq ~ \sum_{k=1}^K \mathbb{E}_{\boldsymbol \epsilon} ~ \sum_{i,k=1}^n \epsilon_{ik}\epsilon_{jk} \big[\langle \phi(\xx_i), \phi(\xx_j) \rangle\big] \\
             & = ~    K \sum_{i=1}^n \langle \phi(\xx_i), \phi(\xx_i) \rangle.
        \end{aligned}
    \end{equation}
    The last step is due to the symmetry of $\langle \phi(\xx_i), \phi(\xx_i) \rangle$ shown in (\ref{equation.non-startionary-rff}).
    The result is similar to \cite{bartlett2002rademacher,li2018multi}
    Applying spectral representation in (\ref{equation.psd-kernel}) of non-stationary kernels, we further bound the Rademacher complexity
    \begin{equation}
        \label{inequation.grc-bound-proof-rff}
        \begin{aligned}
            \widehat{\mathcal{R}}(\mathcal{H})
            \leq ~ & \frac{B}{n} \sqrt{K \sum_{i=1}^n \langle \phi(\xx_i), \phi(\xx_i) \rangle}                             \\
            = ~    & \frac{B}{n} \sqrt{ \frac{K}{D} \sum_{i=1}^n \frac{1}{2}\Big[\cos\big((\OO-\OO')^T \xx_i\big) + 1\Big]}
        \end{aligned}
    \end{equation}
    where $B = \sup_{f \in \mathcal{H}} \|\WW\|_*$.
    Substituting the above inequation (\ref{inequation.grc-bound-proof-rff})  to (\ref{inequation.grc-bound-empirical-grc}), we complete the proof.
\end{proof}

\subsection{Singular Values Thresholding (SVT)}
In each iteration, to obtain a tight surrogate of Equation (\ref{equation.primal-objective}), we keep $\|\WW\|_*$ while relaxing empirical loss $g(\WW)$ only, that leads proximal gradient \cite{parikh2014proximal}
\begin{align*}
    \begin{split}
        &\WW^{t+1}
        =\argmin_{\WW} \lambda_1 \|\WW\|_* + g(\WW)\\
        =&\argmin_{\WW} \lambda_1 \|\WW\|_* + g(\WW^{t})\\
        +&\langle\nabla g(\WW^{t}),\WW-\WW^{t} \rangle
        +\frac{1}{2\eta}\|\WW-\WW^{t}\|_F^2\\
        =&\argmin_{\WW} \lambda_1 \|\WW\|_* + \frac{1}{2\eta}\|\WW-(\WW^{t}-\eta\nabla g(\WW^{t})\|_F^2\\
        =&\argmin_{\WW} \frac{1}{2}\|\WW-{\boldsymbol Q}\|_F^2 + \lambda_1\eta \|\WW\|_*
    \end{split}
\end{align*}
where ${\boldsymbol Q} = \WW^{t}-\eta\nabla g(\WW^{t})$ and $\eta$ is the learning rate.
\begin{proposition}[Theorem 2.1 of \cite{cai2010singular}]
    \label{prop.svt}
    Let ${\boldsymbol Q}\in\mathbb{R}^{D \times K}$ with rank $r$ and its singular values decomposition (SVD) is ${\boldsymbol Q}={\boldsymbol U}{\boldsymbol \Sigma}{\boldsymbol V}^T$,
    where ${\boldsymbol U} \in \mathbb{R}^{d \times r}$ and ${\boldsymbol V} \in \mathbb{R}^{K \times r}$ have orthogonal columns, ${\boldsymbol \Sigma}$ is the diagonal ${\text{diag}(\{\sigma_i\}_{1 \leq i \leq r})}$. Then,
    \begin{align*}
        \argmin_{\WW}\left\{\frac{1}{2}\|\WW-{\boldsymbol Q}\|_F^2 + \eta\|\WW\|_*\right\}={\boldsymbol U}{\boldsymbol \Sigma}_\eta{\boldsymbol V}^T,
    \end{align*}
    where the diagonal is ${\boldsymbol \Sigma}_\eta = \text{diag}(\{\sigma_i - \eta\}_+)$.
\end{proposition}
Applying Proposition \ref{prop.svt} \cite{cai2010singular,lu2015generalized,chatterjee2015matrix} to (\ref{equation.svt_result}), we update $\WW$ twice in each iteration, as shown in \eqref{gradient.W} and \eqref{equation.svt_result}.

\section{Acknowledgments}
This work was supported in part by the National Natural Science Foundation of China (No.61703396, No.61673293), the CCF-Tencent Open Fund,
the Youth Innovation Promotion Association CAS, the Excellent Talent Introduction of Institute of Information Engineering of CAS (No. Y7Z0111107),
and the Beijing Municipal Science and Technology Project (No. Z191100007119002).

\bibliography{3834-References}
\bibliographystyle{aaai}

\end{document}